# HYBRID DEEP CONVOLUTIONAL NEURAL NETWORKS COMBINED WITH AUTOENCODERS AND AUGMENTED DATA TO PREDICT THE LOOK-UP TABLE 2006

Messaoud Djeddou[1]*, Aouatef Hellal[2], Ibrahim A. Hameed[3],

Xingang Zhao[4], and Djehad Al Dallal[5]

[1]Hydraulic Development and Environment Laboratory (LAHE), Faculty of Sciences and Technology, Mohamed Khider University, Biskra, Algeria;

[2]Research Laboratory in Subterranean and Surface Hydraulics (LARHYSS), Faculty of Sciences and Technology, Mohamed Khider University, Biskra, Algeria; email: *hellal.aouatef@gmail.com*

[3]Department of ICT and Natural Sciences, Faculty of Information Technology and Electrical Engineering, Norwegian University of Science and Technology (NTNU), Ålesund, Norway; email: *ibib@ntnu.no*

[4]Nuclear Energy and Fuel Cycle Division, Oak Ridge National Laboratory, Oak Ridge, TN, USA; email: *zhaox2@ornl.gov*

[5]Department of Computer Science, Gulf University for Science & Technology (GUST), West Mishref, Kuwait;

email: *jaldallal@yahoo.com*

*Corresponding author Address*

*messaoud.djeddou@univ-oeb.dz*;

**ABSTRACT**

*This study explores the development of a hybrid deep convolutional neural network (DCNN) model enhanced by autoencoders and data augmentation techniques to predict critical heat flux (CHF) with high accuracy. By augmenting the original input features using three different autoencoder configurations, the model's predictive capabilities were significantly improved.*

*The hybrid models were trained and tested on a dataset of 7225 samples, with performance metrics including the coefficient of determination ($R^2$), Nash-Sutcliffe efficiency (NSE), mean absolute error (MAE), and normalized root-mean-squared error (NRMSE) used for evaluation. Among the tested models, the DCNN_3F-A2 configuration demonstrated the highest accuracy, achieving an $R^2$ of 0.9908 during training and 0.9826 during testing, outperforming the base model and other augmented versions.*

*These results suggest that the proposed hybrid approach, combining deep learning with feature augmentation, offers a robust solution for CHF prediction, with the potential to generalize across a wider range of conditions.*

***Keywords:*** Critical Heat Flux, Deep Convolutional Neural Networks, Auto-encoder, CHF look-up table.

**NOMENCLATURE**

AI     Artificial Intelligence
CHF   Critical Heat Flux
R2     Coefficient of Determination
NSE   Nash-Sutcliffe efficiency
MAE   Mean Absolute Error
RMSE Root-Mean-Squared Error
DCNN Deep Convolutional Neural Network
AE     Auto-Encoders
LUT   Look Up Table

## 1. INTRODUCTION

The Critical Heat Flux (CHF) is the threshold at which the heat transfer between a heated surface and a surrounding fluid reaches its maximum capability, resulting in a significant decrease in heat transfer efficiency. Exceeding this limit in the context of nuclear reactors can lead to a range of unwanted outcomes, including thermal excursion, material degradation, and potentially deadly reactor instabilities. Therefore, it is crucial to comprehend the underlying concepts that govern CHF (Critical Heat Flux) in order to mitigate operating risks and optimize reactor performance. CHF, is a complicated phenomenon influenced by various factors including flow rate, pressure, quality, geometry, and surface qualities [1].

The prediction of the CHF phenomenon has attracted significant investigation from researchers due to its importance in nuclear reactor technology. There are three main types of approaches for CHF predicting:

- Empirical CHF correlations:
- Analytical CHF models;
- CHF look-up table methods.

Every approach has its advantages, but still comes with certain disadvantages such high nonlinearity, complex phenomenology, and uncertainty associated with CHF [2].

The CHF look-up table is essentially a normalized database that includes CHF values predicted based on several factors, including mass flux (G), coolant pressure (P), and thermodynamic quality (x_e_out).

Currently, nuclear reactor thermal hydraulics codes like RELAP5 3D, COBRA-TF, and TRACE use empirical correlations or tabulated look-up tables based on vast experimental data across different operating situations to tackle CHF issues [3][4].

The remarkable advancement in artificial intelligence (AI) in the past three decades has led to an increasing interest in using Ml-based machine learning models for the prediction of critical Heat Flux (CHF) since the 1990s. Several machine learning-based studies have been proposed to predict the CHF look-up table, such as references [5, 6, 7, 8, 9, 10]. Applying a single prediction model may entail some limitations in terms of accuracy. On the other hand, it has been demonstrated that the combination of prediction capabilities from many methodologies enhances the accuracy of predictions, as evidenced by studies [11, 12, 13, 14].

This paper presents a new hybrid model based on auto-encoders (AE) as feature fusion techniques refer to procedures that integrate variables to eliminate redundant and useless information coupled with deep convolutional neural networks (DCNN), for the prediction task of CHF look-up table.

## 2. MATERIALS AND METHODS
### 2.1 Lookup Table
The 2006 Groeneveld Lookup table (LUT) is a commonly used model for estimating the margins of CHF (Critical Heat Flux) in a nuclear reactor. The LUT serves as a tool to forecast CHF in a vast range of variables and is a data-derived estimation of the CHF phenomenon. It can be defined as a standardized database for vertical 8 mm water-cooled tubes [14]. The LUT will serve as a benchmark for evaluating the outcomes of our prediction techniques.

The LUT 2006 employs pressure, mass flux, and quality as the primary parameters, as they are widely recognized as the most influential factors. Furthermore, a correction term is defined for the tube's diameter as stated in reference [14]:

$$CHF = CHF_{8mm} \times \left[\frac{D_{real}}{8}\right]^{0.5} \qquad (1)$$

$CHF_{8mm}$ represents the normalized CHF value obtained from the lookup table, specifically for a diameter of 8mm. Dreal refers to the actual diameter in millimeters. When retrieving $CHF_{8mm}$ from the LUT, it is necessary to do interpolation because to the restricted data points in the table. This interpolation is accomplished using a linear interpolator. The author considered the length of the tube to be a second-order parameter and hence did not include it in the approximation, as long as the L/D ratio was sufficiently large.

### 2.2 Data description
The entire database was built based on the Groeneveld CHF LUT 2006 using 7245 data entries. The training/test partition typically involves randomly partitioning the data into a training set used for models training and a test set used to validate the proposed models. In this study, 80% of the data are used as the training set and 20% as the test set. The input and output variables considered in the study and their descriptive statistics are reported in Table 1.

The training dataset with the three input attributes listed in Table 1 is prepared and Standardization centers data around a mean of zero and a standard deviation of one.

**Table 1.** STATISTICAL PARAMETERS OF THE DATASET.

| Inputs | Min | Max. | Average | S.D. | CV |
|---|---|---|---|---|---|
| Pressure [MPa] | 0.100 | 21 | 8.660 | 7.414 | 0.856 |
| Mass_flux [kg/m$^2$-s] | 0.000 | 8000 | 3295.238 | 2642.460 | 0.802 |
| x_e_out [-] | -0.500 | 1 | 0.220 | 0.403 | 1.834 |
| *Output* | | | | | |
| CHF [MW/m$^2$] | 0.000 | 39744 | 3868.244 | 5419.593 | 1.401 |

## 2.3 Auto-encoders

The concept of autoencoders was initially described in [15] as a type of neural network that is trained to accurately recreate its input. The primary objective is to acquire knowledge through self-directed methods, resulting in a "informative" presentation of the data that can be used to diverse outcomes.

Auto-encodeurs structure is composed of two interconnected networks:

1. Encoder network: converts the initial input, with a high number of dimensions, into a latent code with a lower number of dimensions. The size of the input is greater than the size of the output.
2. Decoder network: The decoder network extracts the information from the code, often by using progressively larger output layers.

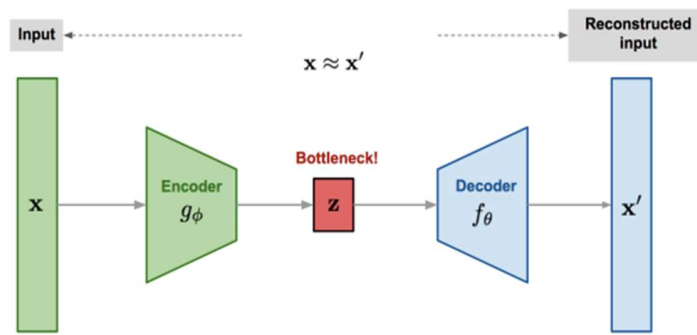

**FIGURE 1:** AUTOENCODER MODELSTRUCTURE.

The model includes an encoding function *g(.)* that is parameterized by ϕ, and a decoding function *f(.)* that is parameterized by θ. The low-dimensional code obtained from input x in the bottleneck layer is denoted as $z = g_\phi(x)$, z=g(x), and the reconstructed input is represented as $x' = f_\theta(g_\phi(x))$.

The parameters (θ,ϕ) are jointly trained to provide a reconstructed data sample that matches the original input, $x \approx f_\theta(g_\phi(x))$, or in simpler terms, to learn an identity function. The metric applied to measure the variance between original input and the reconstructed is mean squared error (MSE) loss:

$$L_{AE}(\theta, \phi) = \frac{1}{n}\sum_{i=1}^{n}\left(x^{(i)} - f_\theta\left(g_\phi(x^{(i)})\right)\right)^2 \qquad (2)$$

### 2.4 Deep Convolutional Neural Networks

Convolutional neural networks, as defined in [16], are a particular kind of neural network characterized by several layers and a feed-forward configuration. The object is a combination of one or more convolutions arranged in layers. Feasible arrangements comprise of input, output, and hidden layers. The input and output layers serve distinct purposes, whereas the hidden layer is commonly employed for performing multiplication or dot product operations. One can create many types of layers, including completely connected layers, normalization layers, and pooling layers [17].

A deep convolutional neural network (DCNN) is a specific kind of neural network. network that consists of multiple convolutional layers. This architecture allows the network to efficiently process a vast volume of data and produce accurate results. Please refer to Figure 3 for a visual representation of this architecture. DCNN, with its weight sharing structure and pooling algorithms, effectively reduces the parameter count and surpasses DNNs, especially in the analysis of visual pictures. CNN, or Convolutional Neural Network, is spatially invariant, meaning it does not encode an object's location and orientation. If the precise positioning of data is crucial, the application of CNN may not be straightforward. Convolutional Neural Networks (CNNs) are becoming more used in a range of water and wastewater treatment applications.

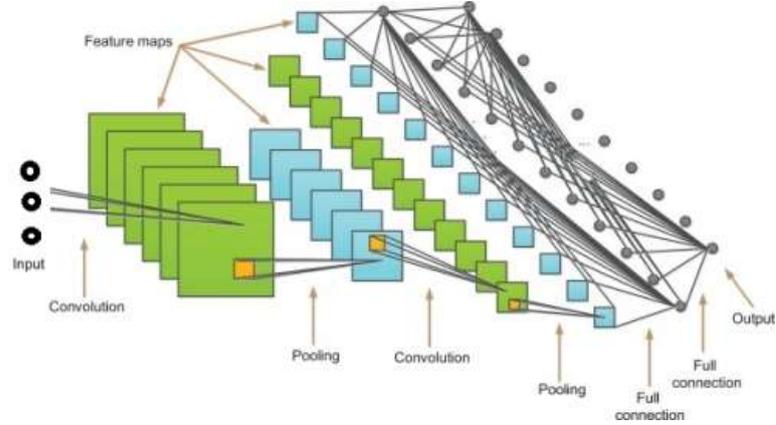

**FIGURE 2:** DEEP CONVOLUTIONAL NEURAL NETWORKS.

## 2.5 Deep Performance evaluation

Training and testing process of DCNN models were assessed using statistical parameters such as the normalized root mean square error (NRMSE), mean absolute error (MAE), coefficient of determination ($R^2$), and Nash-Sutcliffe efficiency coefficient (NSE). These parameters were expressed as follows:

$$NRMSE = \frac{\sqrt{\frac{\sum_{i=1}^{N}\left(CHF_i^{measured}-CHF_i^{predicted}\right)^2}{N-1}}}{\overline{CHF}} \quad (3)$$

$$MAE = \frac{1}{N}\sum_{i=1}^{N}\left|CHF_i^{measured} - CHF_i^{predicted}\right| \quad (4)$$

$$NSE = 1 - \frac{\sum_{i=1}^{N}\left(CHF_i^{measured}-CHF_i^{predicted}\right)^2}{\sum_{i=1}^{N}\left(CHF_i^{measured}-\overline{CHF_i^{measured}}\right)^2} \quad (5)$$

$$R^2 = \left[\frac{\sum_{i=1}^{N}\left(CHF_i^{measured}-\overline{CHF_i^{measured}}\right)\left(CHF_i^{predicted}-\overline{CHF_i^{predicted}}\right)}{\sqrt{\sum_{i=1}^{N}\left(CHF_i^{measured}-\overline{CHF_i^{measured}}\right)^2 \sum_{i=1}^{N}\left(CHF_i^{predicted}-\overline{CHF_i^{predicted}}\right)^2}}\right]^2 \quad (6)$$

where: CHF: Critical Heat Flux (MW/m$^2$), $\overline{CHF}$: mean CHF (MW/m$^2$).

## 3. RESULTS AND DISCUSSION

The DCNN model consisted of 5 convolutional layers, with a similar outcome where higher layers did not significantly improve loss. The training and testing were carried out on a computer equipped with an ASUS AMD Rayzen TM 5 R5-3550H CPU at 3.7 GHz and 16 GB of RAM.

The use of 3 different auto-encoders for features augmentation led to the increase of features as represented in Table 2.

**TABLE 2:** FEATURES AUGMENTATION USING DIFFERENT DECOMPOSITION TECHNIQUES

| Models | Initial features | Final features |
|---|---|---|
| DCNN_3F | 3 | 3 |
| DCNN_3F-A1 | 3 | 4 |
| DCNN_3F-A2 | 3 | 5 |
| DCNN_3F-A3 | 3 | 6 |

Where: F: Feature; A: Augmented feature.

The models were constructed using the Keras library in Python, which provides an interface for Deep Learning. The mean-squared-error (MSE) was selected as the loss function due to its high sensitivity to large errors compared to mean absolute error. The Adam optimizer was utilized to optimize the DCNN models. All models were trained on the same dataset of 7225 samples, with the datasets being randomly divided into training and testing sets in a 80:20 ratio. The hyperparameter values that were selected are presented in Table 3.

**TABLE 3:** DEVELOPED MODELS DESCRIPTION

|  | **DCNN_3F** | **DCNN_3F-A1** | **DCNN_3F-A2** | **DCNN_3F-A3** |
|---|---|---|---|---|
| Input shape | (n, 3, 1) | (n, 4, 1) | (n, 5, 1) | (n, 6, 1) |
| Number of layers | Total of 8 | Total of 8 | Total of 8 | Total of 8 |
| Batch size | 32 | 32 | 32 | 32 |
| Number of epochs | 200 | 200 | 200 | 200 |
| Loss function | MSE | MSE | MSE | MSE |
| Optimizer | Adam | Adam | Adam | Adam |
| Learning rate | $10^{-3}$ | $10^{-3}$ | $10^{-3}$ | $10^{-3}$ |
| Number of parameters | 35329 | 35329 | 35329 | 47617 |

The obtained results of the proposed models using DCNN for the CHF prediction are presented in Table 4. From the overall comparison, it can be observed that all the model combinations demonstrate accurate performance in both train and test phases; this can be proved by considering the $R^2$ values that are greater than 0.80.

**TABLE 4:** PERFORMANCE RESULTS OF RBFNN, GRNN AND MLPNN MODELS.

| Model | Training | | | | Testing | | | |
|---|---|---|---|---|---|---|---|---|
|  | NRMSE | MAE | $R^2$ | NSE | NRMSE | MAE | $R^2$ | NSE |
| *DCNN_3F* | 0.1154 | 385.1688 | 0.9875 | 0.9867 | 0.1478 | 420.7925 | 0.9791 | 0.9781 |
| *DCNN_3F-A1* | 0.1091 | 314.7993 | 0.9882 | 0.9881 | 0.1389 | 344.7399 | 0.9807 | 0.9807 |
| *DCNN_3F-A2* | 0.1003 | 331.6780 | 0.9908 | 0.9899 | 0.1356 | 354.3443 | 0.9826 | 0.9816 |
| *DCNN_3F-A3* | 0.1236 | 425.3256 | 0.9851 | 0.9847 | 0.1870 | 542.8060 | 0.9653 | 0.9650 |

## 3.1 Simple model (DCNN 3F)

Table 4 shows that the simple model in training phase produces an NRMSE value of the simple model is equal to 0.1154 indicates that, on average, the model's predictions deviate by approximately 11.54% from the actual values in the training dataset. With an MAE of 385.17 (MW/m$^2$), the model's average prediction error in the training dataset is around 385 (MW/m$^2$). The $R^2$ value of 0.9875 indicates that the model explains approximately 98.75% of the variance in the training data. With an NSE of 0.9867, the model's performance on the training dataset is quite high.

In testing phase, The NRMSE value of 0.1478 indicates that the model's predictions deviate by approximately 14.78% from the actual values, the MAE value of 420.79 (MW/m$^2$). The $R^2$ value of 0.9791 for the testing dataset indicates that the model explains approximately 97.91% of the variance in the testing data. The NSE value of 0.9781 for the testing dataset indicates high efficiency in predicting the target variable. The model demonstrates strong performance on both the training and testing datasets across multiple evaluation metrics, indicating good generalization ability

The fitting between the measured and predicted CHF is shown in figure 5.

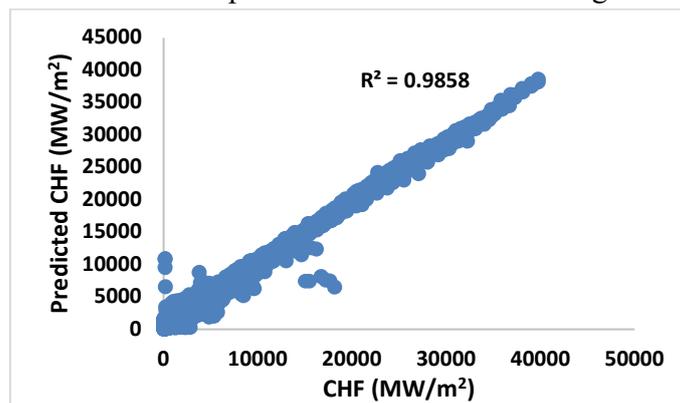

**Figure 7:** MEASURED CHF VS PREDICTED CHF WITH SIMPLE MODEL (DCNN 3F).

## 3.2 Hybrid models

It can be observed from table 4, that using different auto-encoders four feature augmentation can increase the performance outcomes in CHF prediction. Furthermore, an explanation of the results revealed that for predicting the CHF, DCNN_3F-A2 with NRMSE (0.1003), MAE (331.67), $R^2$ (0.9908) and *NSE* (0.9899) values in the training phase, proved merit over DCNN_3F-A1 and DCNN_3F-A3.

In testing phase, the DCNN_3F-A2 with NRMSE (0.1356), MAE (354.344), $R^2$ (0.9826) and *NSE* (0.9816) outperforms the DCNN_3F-A1 and DCNN_3F-A3 and therefore emerged as a reliable model.

Figure 8 show the fitting between the measured and predicted CHF values in the entire dataset.

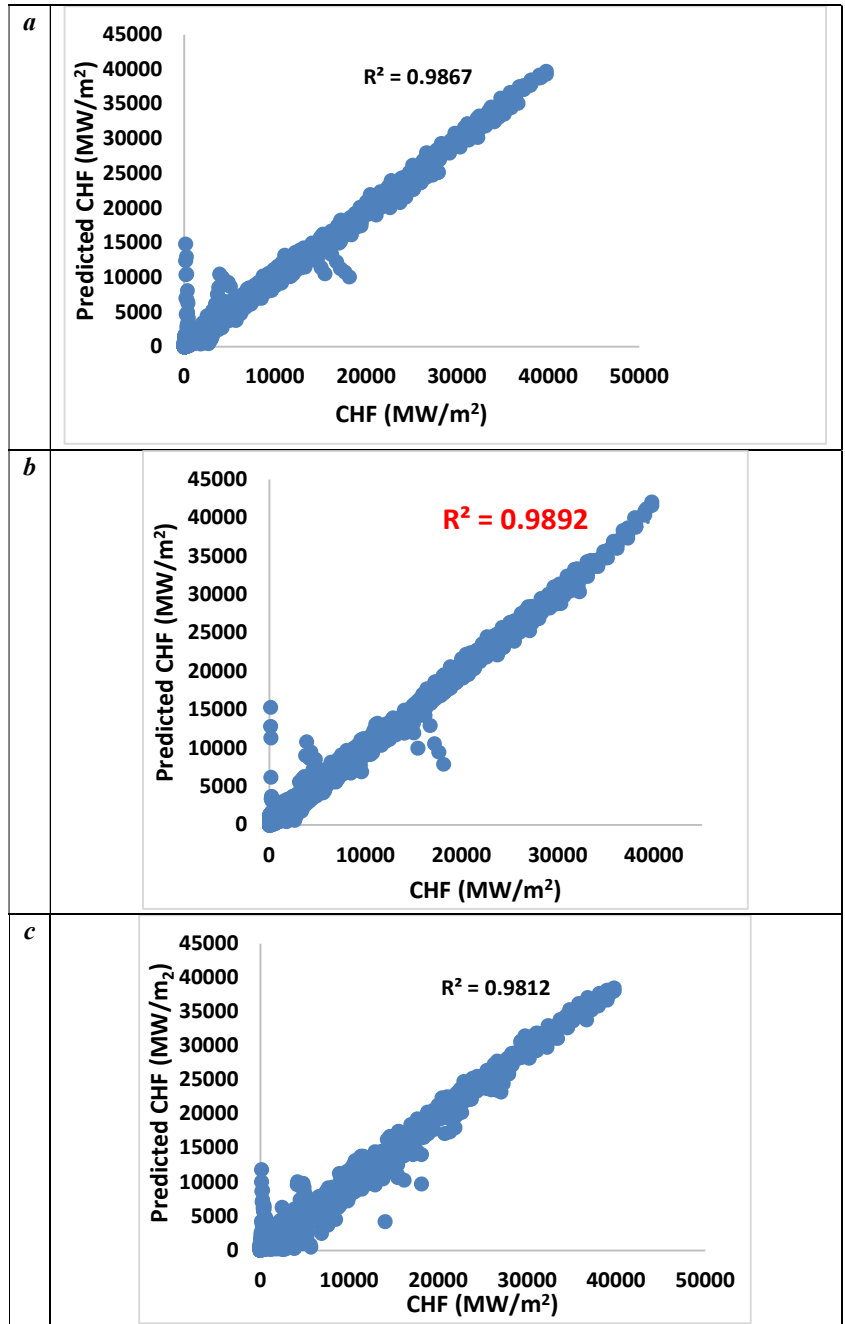

**Figure 8:** MEASURED CHF VS PREDICTED CHF WITH HYBRID MODELS (*a*) DCNN_3F-A1, (*b*) DCNN_3F-A2, (*c*) DCNN_3F-A3).

## 4. CONCLUSION

This research introduced the novel application of a hybrid approach using Auto-encoders (AE) and deep convolutional neural networks (DCNN) for predicting Critical Heat Flux (CHF). Auto-encoders were employed as a technique for enhancing features. The rebuilt features were combined with the main features and utilized as inputs for various hybrid deep convolutional neural networks (DCNN).

The performance trends across the models highlight the importance of carefully selecting the number of augmented features when designing hybrid models. While feature augmentation generally improves model performance, an optimal balance must be struck to avoid overfitting or introducing unnecessary complexity. The DCNN_3F-A2 model's superior performance indicates that two additional features, derived from autoencoders, provided the best enhancement in predictive accuracy without overwhelming the model with excessive information.

In conclusion, the study confirms that hybrid models leveraging feature augmentation through autoencoders can significantly improve the predictive performance of deep learning models in complex tasks such as CHF prediction. However, the effectiveness of this approach is highly dependent on the appropriate selection and number of augmented features.

The suggested AE-DCNN (3F-A2) model serves as a baseline for predictive modeling of Critical Heat Flux (CHF) and can be utilized as a tool in experimental management.

**ACKNOWLEDGEMENTS**
The authors would like to thank:
Mohamed Khider University of Biskra, Algeria, Norwegian University of Science and Technology (NTNU) Norway, and Gulf University for Science and Technology, Kuwait for their support.